\documentclass[11pt,a4paper]{article}
\usepackage[hyperref]{emnlp2020}
\usepackage{times}
\usepackage{latexsym}

\usepackage{booktabs} 
\usepackage{graphicx}
\usepackage{microtype}
\usepackage{amsmath}
\usepackage{amsfonts}       
\usepackage{nicefrac}
\usepackage{multirow}
\usepackage{comment}
\usepackage{enumitem}
\aclfinalcopy 

\title{Cross-Thought for Sentence Encoder Pre-training}

\author{Shuohang Wang$^1$ \quad Yuwei Fang$^1$ \quad Siqi Sun$^1$ \quad Zhe Gan$^1$ \quad Yu Cheng$^1$ \\ \textbf{Jing Jiang}$^2$  \quad \textbf{Jingjing Liu}$^1$\\ \\
     $^1$Microsoft Dynamics 365 AI Research, $^2$Singapore Management University \\
    {\small \{\tt shuowa, yuwfan, siqi.sun, zhe.gan, yu.cheng, jingjl\}@microsoft.com} \\
    \small \{\tt jingjiang\}@smu.edu.com
    }

\date{}

\begin{document}
\maketitle
\begin{abstract}
In this paper, we propose Cross-Thought, a novel approach to pre-training sequence encoder, which is instrumental in building reusable sequence embeddings for large-scale NLP tasks such as question answering. 
Instead of using the original signals of full sentences, we train a Transformer-based sequence encoder over a large set of short sequences, which allows the model to automatically select the most useful information for predicting masked words. 
Experiments on question answering and textual entailment tasks demonstrate that our pre-trained encoder can outperform state-of-the-art encoders trained with continuous sentence signals as well as traditional masked language modeling baselines.
Our proposed approach also achieves new state of the art on HotpotQA (full-wiki setting) by improving intermediate information retrieval performance.\footnote{Our code will be released at \url{https://github.com/shuohangwang/Cross-Thought}.}

\end{abstract}

\section{Introduction}
Encoding sentences into embeddings~\cite{skipthought,subramanian2018learning,sentencebert} is a critical step in many Natural Language Processing (NLP) tasks.
The benefit of using sentence embeddings is that the representations of all the encoded sentences can be reused on a chunk level (compared to word-level embeddings), which can significantly accelerate inference speed.
For example, when used in question answering (QA), it can significantly shorten inference time with all the embeddings of candidate paragraphs  pre-cached into memory and only matched with the question embedding during inference. 

\begin{figure*}[t]
\centering
\includegraphics[width=6in]{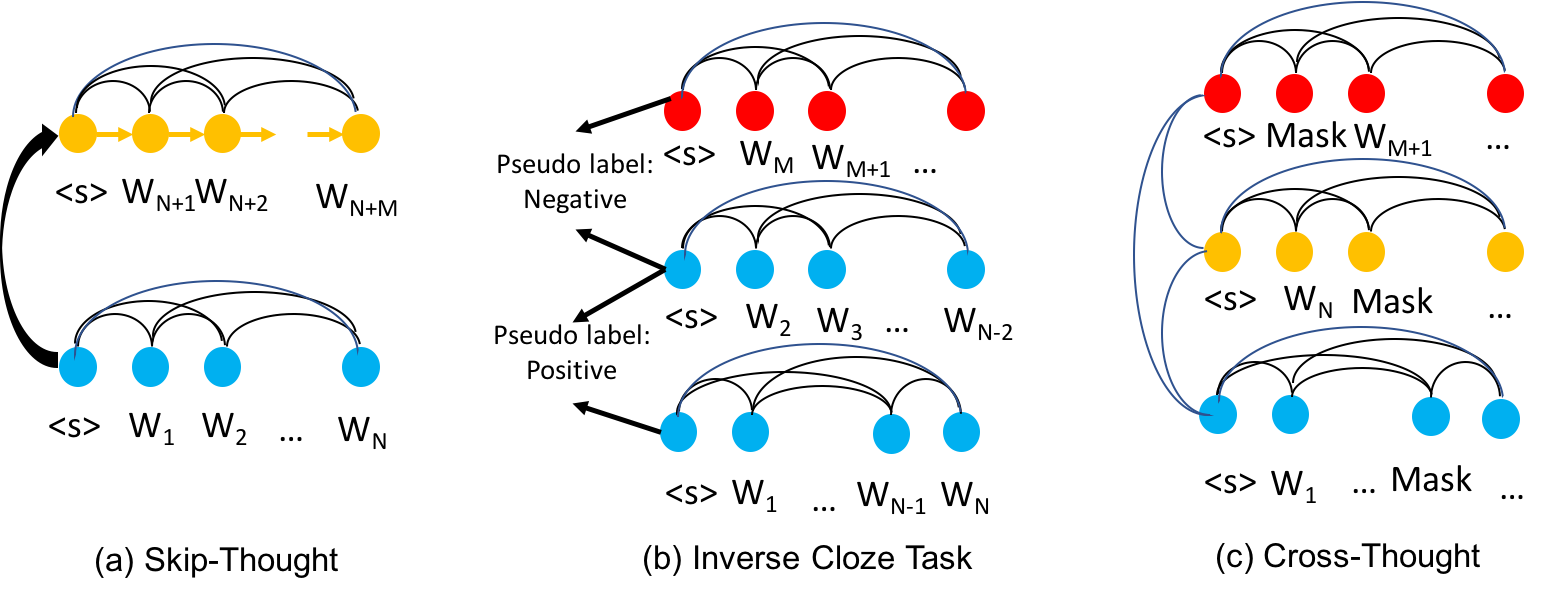}
\caption{Structures of pre-training models for sentence encoder. (a) is a Seq2Seq model that generates the next sentence based on the embedding of the previous sentence. (b) is the classification/ranking model based on pre-defined pseudo labels. (c) is the structure of our model by using the sentence embedding from other sequences to generate the masked word.}
\label{fig:tasks}
\end{figure*}

There have been several models specifically designed to pre-train sentence encoders with large-scale unlabeled corpus.
For example, Skip-thought~\cite{skipthought} uses encoded sentence embeddings to generate the next sentence~(Figure~\ref{fig:tasks}(a)). 
Inverse Cloze Task~\cite{ict} defines some pseudo labels to pre-train a sentence encoder~(Figure~\ref{fig:tasks}(b)).
However, pseudo labels may bear low accuracy, and rich linguistic information that can be well learned in generic language modeling is often lost in these unsupervised methods.
In this paper, we propose a novel unsupervised approach that fully exploits the strength of language modeling for sentence encoder pre-training.

\begin{figure}[t]
\centering
\includegraphics[width=3in]{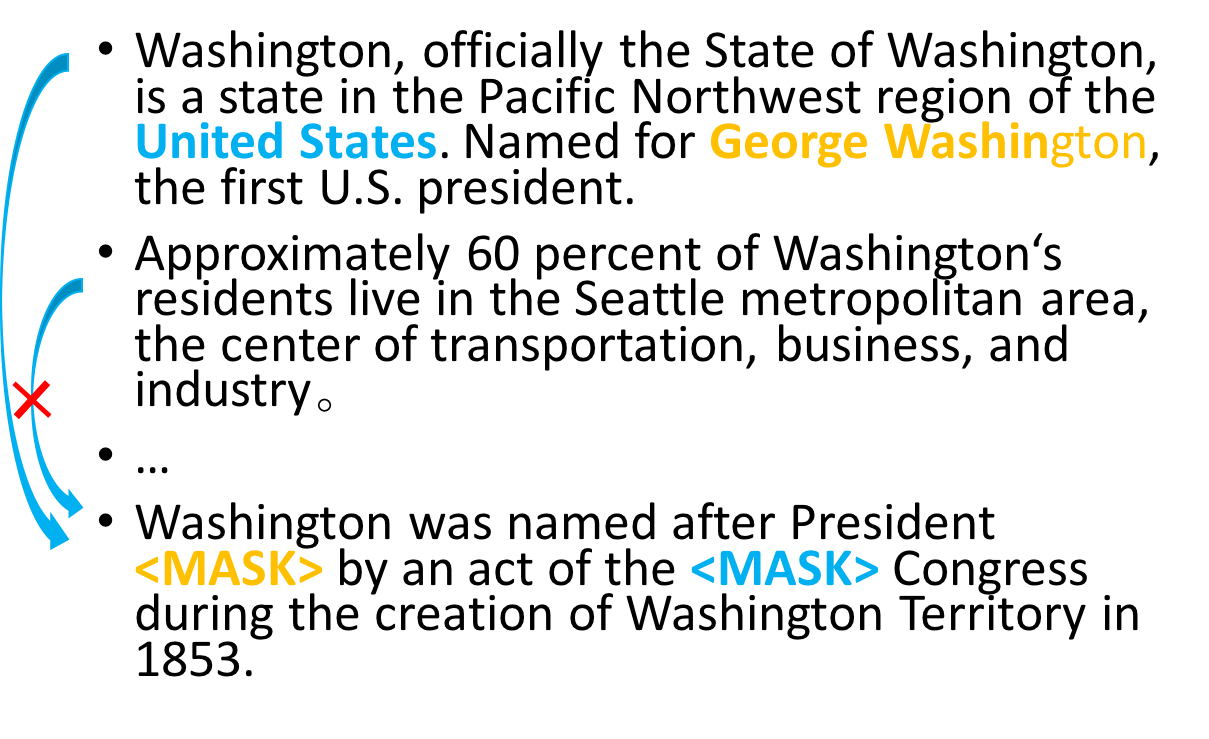}
\vspace{-1cm}
\caption{Example of short sequences that can leverage each other for pre-training sentence encoder.}
\label{fig:example}
\end{figure}

Popular pre-training tasks such as language modeling~\cite{elmo,gpt}, masked language modeling~\cite{bert,roberta} and sequence generation~\cite{unilm,bart} 
are not directly applicable to sentence encoder training, because only the hidden state of the first token (a special token)~\cite{sentencebert,bert} can be used as the sentence embedding, but no loss or gradient is specifically designed for the first special token, which renders sentence embeddings learned in such settings contain limited useful information.

Another limitation in existing masked language modeling methods~\cite{bert,roberta} is that they focus on long sequences (512 words), where masked tokens can be recovered by considering the context within the same sequence. This is useful for word dependency learning within a sequence, but less effective for sentence embedding learning.

In this paper, we propose Cross-Thought, which segments input text into shorter sequences, where masked words in one sequence are less likely to be recovered based on the current sequence itself, but more relying on the embeddings of other surrounding sequences. For example, in Figure~\ref{fig:example}, the masked words ``George Washington" and ``United States" in the third sequence can only be correctly predicted by considering the context from the first sequence. Thus, instead of performing self-attention over all the words in all sentences, our proposed pre-training method enforces the model to learn from mutually-relevant sequences and automatically select the most relevant neighbors for masked words recovery.



The proposed Cross-Thought architecture is illustrated in Figure~\ref{fig:tasks}(c). Specifically, we pre-append each sequence with multiple special tokens, the final hidden states of which are used as the final sentence embedding.
Then, we train multiple cross-sequence Transformers over the hidden states of different special tokens independently, to retrieve relevant sequence embeddings for masked words prediction.
After pre-training, the attention weights in the cross-sequence Transformers can be directly applied to downstream tasks (e.g., in QA tasks, similarity scores between question and candidate answers can be ranked by their respective sentence embeddings).

Our contributions are summarized as follows.
$(i)$ We propose the Cross-Thought model to pre-train a sentence encoder with a novel pre-training task: recovering a masked short sequence by taking into consideration the embeddings of surrounding sequences.
$(ii)$ Our model can be easily finetuned on diverse downstream tasks. The attention weights of the pre-trained cross-sequence Transformers can also be directly used for ranking tasks.
$(iii)$ Our model achieves the best performance on multiple sequence-pair classification and answer-selection tasks, compared to state-of-the-art baselines. In addition, it further boosts the recall of information retrieval (IR) models in open-domain QA task, and achieves new state of the art on the HotpotQA benchmark (full-wiki setting).

\section{Related Work}
\paragraph{Sequence Encoder} 
Many studies have explored different ways to improve sequence embeddings. 
\citet{huang2013learning} proposes deep structured semantic encoders for web search.
\citet{tan2015lstm} uses LSTM as the encoder for non-factoid answer selection, and \citet{tai2015improved} proposes tree-LSTM to compute semantic relatedness between sentences.
\citet{mou2015:emnlp} also uses tree-based CNN as the encoder for textual entailment tasks.
 \citet{cheng2016long} proposes Long Short-Term Memory-Networks (LSTMN) for inferring the relation between sentences, and \citet{lin2017structured} combines LSTM and self-attention mechanism to improve sentence embeddings.
Multi-task learning~\cite{subramanian2018learning,cer2018universal} has also been applied for training better sentence embeddings.
Recently, in additional to supervised learning, models pre-trained with unsupervised methods begin to dominate the field. 

\paragraph{Pre-training}
Several methods have been proposed to directly pre-train sentence embedding, such as Skip-thought~\cite{skipthought}, FastSent~\cite{hill2016learning}, and Inverse Cloze Task~\cite{ict}.
Although these methods can obtain better sentence embeddings in an unsupervised way, they cannot achieve state-of-the-art performance in downstream tasks even with further finetuning.
More recently, \citet{elmo} proposes to pre-train LSTM with language modeling (LM) task, and \citet{gpt} pre-trains Transformer also with LM.
Instead of sequentially generating words in a single direction, \citet{bert} proposes the masked language modeling task to pre-train bidirectional Transformer.
Most recently, \citet{realm,lewis2020pre} propose to jointly train sentence-embedding-based information retriever and Transformer to re-construct documents.
However, their methods are usually difficult to train  with reinforcement learning methods involved, and need to periodically re-index the whole corpus such as Wikipedia. 
In this paper, to pre-train sentence encoder, we propose a new model Cross-Thought to recover the masked information across sequences. We make use of the heuristics that nearby sequences in the document contain the most important information to recover the masked words. Therefore, the challenging retrieval part can be replaced by soft-attention mechanism, making our model much easier to train. 
\section{Cross-Thought}

\begin{figure*}[t!]
\centering
\includegraphics[width=6.3in]{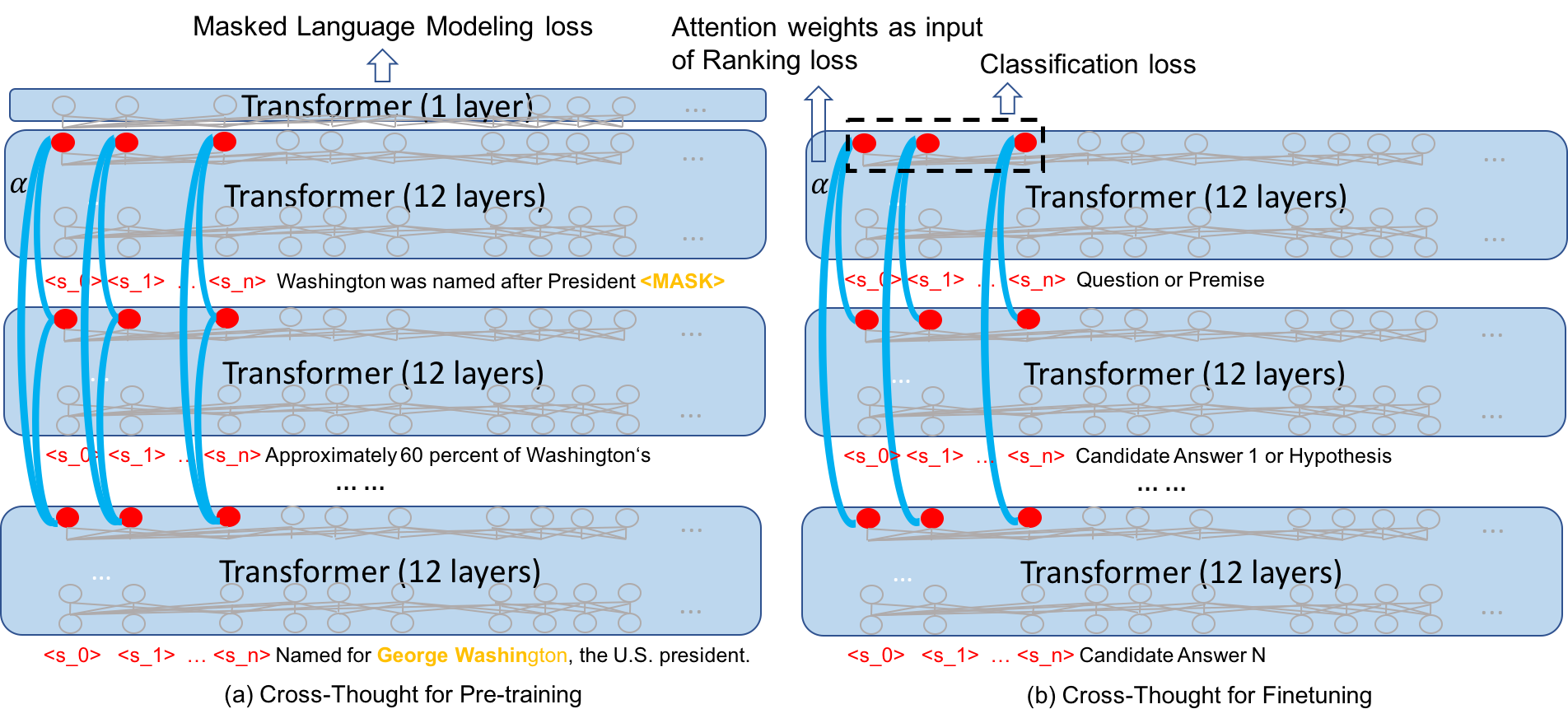}
\caption{Illustration of Cross-Thought for pre-training and finetuning procedures. Circles in red are sentence embeddings. Lines in blue are the cross-sequence Transformers, the attention weights of which are $\alpha$ in Eqn.~(\ref{eqn:att}). Words in red are special tokens, the hidden states of which are the sentence embeddings. Multiple special tokens are used to enrich the sentence embeddings. In (a), the sentence embedding of the third sequence provides context that can help generate the masked word in the first sequence. In (b), the model can be initialized with pre-trained Cross-Thought for Answer Selection or Textual Entailment tasks. The attention weights $\alpha$ and hidden states of cross-sequence Transformers can be used directly for Ranking and Classification tasks. }
\label{fig:model}
\end{figure*}

In this section, we introduce our proposed  pre-training model Cross-Thought, and describe how to finetune the pre-trained model on downstream tasks. 
Specifically, most parameters in downstream tasks can be initialized by the pre-trained Cross-Thought, and for certain tasks (e.g., ranking) the attention weights across sequences can be directly used without additional parameters~(Figure~\ref{fig:model}).

\subsection{Pre-training Data Construction}

Our pre-training task is inspired by Masked Language Modeling~\cite{bert,roberta}, and the key difference is the way to construct sequences for pre-training.
As our goal is sentence embedding learning, the pre-training task is designed to encourage the model to recover masked words based on sentence-level global context from other sequences, instead of word-level local context within the same sequence (Figure~\ref{fig:example}).
Therefore, unlike previous work that segments raw text into long sequences and shuffles the sequences for pre-training, we propose to create shorter text sequences instead, without shuffling.
In this way, a shorter sequence may not contain all the necessary information for recovering the masked words, hence requiring the probing into surrounding sequences to capture the missing information.

\subsection{Cross-Thought Pre-training}
The pre-training model is illustrated in Figure \ref{fig:model}(a).
As aforementioned, the input of pre-training data consists of $M$ continuous sequences $[X_0,X_1, ..., X_{M-1}]$.
Similar to BERT~\cite{bert}, we use the hidden state of the special token as the final sentence embedding. 
To encode the embeddings with richer semantic information, we propose to pre-append $N$ special tokens $S$ instead of a single one to each sequence $X$.

We first use Transformer to encode the segmented short sequences as follows:
\begin{eqnarray}
\mathbf{H}_m &=& \text{Transformer}([\mathbf{S}; \mathbf{X}_m]), \label{eqn:trans}  
\\
\mathbf{E}_m &=& \mathbf{H}_{m}[0{:}N-1],
\end{eqnarray}
where $\mathbf{S}\in \mathbb{R}^{N\times d}$ are the embeddings of $N$ special tokens, and $\mathbf{X}_m \in \mathbb{R}^{l_m\times d}$ are the contextualized word embeddings of the $m$-th sequence. 
$l_m$ is the sequence length and $d$ is the dimension of the embedding.
$[\cdot;\cdot]$ is the concatenation of matrices.
$\mathbf{H}_{m}\in \mathbb{R}^{(N+l_m)\times d}$ are all the hidden states of the Transformer, and $\mathbf{E}_m\in \mathbb{R}^{N\times d}$ are the hidden states on the special tokens, used as the final sequence embedding.

Next, we build cross-sequence Transformer on top of these sequence embeddings, so that each sequence can distill information from other sequences.
As the embeddings of different special tokens encode different information, we run Transformer on the embedding of each special token separately:
\begin{eqnarray}
\mathbf{F}_{n} &=& \left[\mathbf{E}_{0}[n]; \mathbf{E}_{1}[n];...;\mathbf{E}_{M-1}[n]\right], \\
\mathbf{C}_{n} &=& \text{Cross-Transformer}\left( \mathbf{F}_{n} \right),
\label{eqn:crotrans}    
\end{eqnarray}
where $\mathbf{E}_{m}[n]\in \mathbb{R}^d$ is the $n$-th row of $\mathbf{E}_{m}$.
$\mathbf{F}_{n}\in \mathbb{R}^{M\times d}$ is the concatenation of all the embeddings of the $n$-th special token in each sequence.
$\mathbf{C}_{n}\in \mathbb{R}^{M\times d}$ is the output of cross-sequence Transformer, where all the information across sequences are fused together.
As the weights of multi-head attention in cross-sequence Transformer will be used for downstream tasks, we decompose the attention weights of one head in the cross-sequence Transformer on the $n$-th special tokens as follows:
\begin{eqnarray}
    \mathbf{Q} &=& \mathbf{W}^Q  \mathbf{F}_{n}, \\
    \mathbf{K} &=& \mathbf{W}^K \mathbf{F}_{n}, \\
    \mathbf{\alpha} &=& \text{ Softmax } ( \frac{\mathbf{Q}\mathbf{K}^T}{ \sqrt{d} }),
    \label{eqn:att}
\end{eqnarray}   
where $\mathbf{W}^Q\in \mathbb{R}^{d\times d},\mathbf{W}^K\in \mathbb{R}^{d\times d}$ are the parameters to learn.
$\mathbf{\alpha}\in \mathbb{R}^{M\times M}$ are the attention weights that can be directly used in downstream tasks (e.g., for measuring the similarity between question and candidate answers in QA tasks).

Finally, to encourage the embedding from other sequences retrieved by cross-sequence Transformer to help generate the masked words in the current sequence, we use another Transformer layer on top of the merged sequence embeddings  as follows:
\begin{eqnarray}
\nonumber
\mathbf{G}_m &=& \left[\mathbf{C}_0[m];\mathbf{C}_1[m];...;\mathbf{C}_{N-1}[m];\mathbf{H}_m[N{:}]\right], \\
\mathbf{O}_m &=& \text{Transformer}(\mathbf{G}_m) , 
\end{eqnarray}
where $\mathbf{C}_n[m]\in \mathbb{R}^d$ is the hidden state of cross-sequence Transformer on the $n$-th special token of the $m$-th sequence.
$\mathbf{H}_m[N{:}]\in \mathbb{R}^{l_m\times d}$ from Eqn.(\ref{eqn:trans}) is the hidden state for non-special words $X_m$.
$\mathbf{O}_m\in \mathbb{R}^{(N+l_m)\times d}$ will be used to generate the masked word:
\begin{eqnarray}
\mathcal{L}_{\text{mask}} &=& \sum_{m,i} -\log( P(a_{m,i}|\mathbf{O}_m) ), 
\end{eqnarray}
where $P(a_{m,i}|\mathbf{O}_m)$ is the probability of generating the $i$-th masked word in the $m$-th sequence.

\subsection{Cross-Thought Finetuning}
To demonstrate how pre-trained Cross-Thought can initialize models for downstream tasks, we take two tasks as examples: answer selection and sequence-pair classification (under the setting of using sentence embeddings only, without word-level cross-sequence attention~\cite{bert}).
The procedure is illustrated in Figure~\ref{fig:model}~(b).

\paragraph{Answer Selection} The goal is to select one answer from a candidate pool, $\{X_1, X_2, ..., X_{M-1} \}$ based on question $X_0$.
We consider the representations of candidate answers that are cached and can be matched to question embeddings when a new question comes in.
Based on the pre-trained model, the attention weights in Eqn.(\ref{eqn:att}) from different heads of the cross-sequence Transformers can be directly applied to rank the answer candidates.
For further finetuning, the loss relying on the attention weights is defined as follows:
\begin{eqnarray}
    \mathcal{L}_{\text{answer}} = -\log(\mathbf{\alpha}[0][m]), 
\end{eqnarray}    
where $\mathbf{\alpha}[0]$ are the attention weights between question $X_0$ and all the answer candidates, and $m$ is the index of the correct answer in $\{X_1, X_2, ..., X_{M-1} \}$.
Note that we have multiple cross-sequence Transformers on different special tokens, and each Transformer has multiple heads. Thus, we use the mean value of all the attention matrices as the final weights.

\paragraph{Sentence Pair Classification} The goal is to identify the relation between two sequences, $X_0$ and $X_1$.
Reusable sequence embeddings are very useful in some tasks, such as finding the most similar pair of sentences from a large candidate pool, which requires large-scale repetitive encoding and matching without pre-computed sentence embeddings.
As the pre-training of cross-sequence Transformer is designed to fuse the embeddings of different sequences, the merged representations in Eqn.(\ref{eqn:crotrans}) can be used for downstream classification as follows:
\begin{eqnarray}
    \nonumber
    \mathbf{\bar{C}} &=& [\mathbf{C}_0; \mathbf{C}_1; ...; \mathbf{C}_{N-1} ], \\
    \nonumber
    \mathbf{\bar{c}} &=& \text{Flatten}(\mathbf{\bar{C}}), \\
    \mathcal{L}_{\text{cls}} &=& -\log \left( \text{Softmax}\left(\mathbf{W}^c \mathbf{\bar{c}}^T\right)[y]\right),
\end{eqnarray} 
where $\mathbf{\bar{C}}\in \mathbb{R}^{2N\times d}$ is the concatenation of the hidden states of all the cross-sequence Transformers on $N$ different special tokens. 
Note that there are only two sequences here for classification.
$\mathbf{\bar{c}}\in \mathbb{R}^{2Nd}$ is the reshaped matrix for final classification, and cross-entropy loss is used for optimization.
\section{Experiments}

In this section, we conduct experiments based on our pre-trained models, and provide additional detailed analysis.
\begin{table}[t!]
\resizebox{\linewidth}{!}{
\begin{tabular}{lcccc}
\toprule
Dataset  & \#train & \#test & \#seq & Goal     \\
\midrule
MNLI-m   & 373K    & 10K    & 2     & classification \\
MNLI-mm  & 373K    & 10K    & 2     & classification \\
SNLI     & 549K    & 10K    & 2     & classification \\
QQP      & 346K    & 391K   & 2     & classification \\
Quasar-T & 29K     & 3K     & 100   & ranking     \\
HotpotQA & 86K     & 7K     & 5M    & ranking    \\
\bottomrule
\end{tabular}
}
\caption{Statistics of the datasets. \#train and \#test are the number of samples for training and testing. \#seq is the number of sequences needed to use for each sample. 5M is for 5 million.  }
\label{tbl:stat}
\end{table}

\begin{table*}[t]
\centering
\addtolength{\tabcolsep}{-2pt}
\begin{tabular}{lccccccc}
\toprule
               & MNLI-m & MNLI-mm & SNLI & QQP  & Quasar-T & HotpotQA & HotpotQA(u) \\
               & Acc & Acc & Acc & Acc  & Recall@1 & Recall@20 & Recall@20 \\
               \midrule
ICT            & 65.2   & 65.3    & 81.8 & 85.0 & 40.5     & 81.5     & 44.9        \\
Skip-Thought & 65.3   & 65.7    & 81.5 & 84.6 & 40.2     & 82.0     & 37.6           \\
Language Model  & 73.9   & 74.2    & 84.1 & 86.8 & 41.9     & 81.0     & 33.4         \\
Masked LM-1-512  & 74.3   & 74.1    & 84.9 & 87.3 & 43.2     & 81.5     & 10.0        \\
Masked LM-1-64   & 73.8   & 74.0    & 84.3 & 87.0 & 42.5     & 81.0     & 10.0        \\
Masked LM (160G)  & 75.5   & 75.7    & 86.3 & 89.3 & 43.5     & 87.5     & 10.0        \\
\midrule
Cross-Thought-1-512           & 74.5   & 74.1    & 85.0 & 87.5 & 43.5     & 81.7     & 10.0        \\
Cross-Thought-1-64            & 76.2   & 76.4    & 86.3 & 90.0 & 47.2     & 88.0     & 51.9        \\
Cross-Thought-3-64            & 76.5   & \textbf{76.6}    & 86.5 & \textbf{90.3} & 48.2     & 88.4     & 55.4        \\
Cross-Thought-5-64            & \textbf{76.8}   & \textbf{76.6}    & \textbf{86.8} & \textbf{90.3} & \textbf{48.5}     & \textbf{88.9}     & \textbf{56.5}        \\
\bottomrule
\end{tabular}
\caption{Results on only using sentence embedding for classification and ranking. Cross-Thought-3-64 is to train Cross-Thought by pre-appending 3 special tokens to the sequences that are segmented into 64 tokens. For HotpotQA, we only evaluate on how well the model can retrieve gold paragraphs. Results for HotpotQA(u) are without finetuning. Acc: Accuracy. Recall@20: recall for the top 20 ranked paragraphs.}
\label{tbl:exp}
\end{table*}

\begin{table*}[t]
\centering
\begin{tabular}{lccc}
\toprule
\multicolumn{1}{c}{\multirow{2}{*}{Models}} & \multicolumn{3}{c}{HotpotQA (full-wiki)} \\
\multicolumn{1}{c}{}                        & Pas EM   & Ans EM/F1  & Sup EM/F1  \\
\midrule
Cognitive Graph~\cite{ding2019cognitive}                             & 57.8     & 37.6/49.4    & 23.1/58.5   \\
Semantic Retrieval~\cite{nie2019revealing}                          & 63.9     & 46.5/58.8    & 39.9/71.5   \\
Recurrent Retriever~\cite{asai2019learning}                         & 72.7     & 60.5/73.3    & 49.3/76.1   \\
\midrule
Masked LM-1-64\;+\;reranker\;+\;reader                             & 77.2     & 60.9/73.5    & 52.9/77.3   \\
Cross-Thought-1-64\;+\;reranker\;+\;reader                                   & \textbf{80.0}     & \textbf{62.3}/\textbf{75.1}    & \textbf{54.3}/\textbf{78.6 } \\
\bottomrule
\end{tabular}
\caption{Results on HoptpotQA (full-wiki setting). We use sentence embeddings from the finetuned model of Cross-Thought or Masked LM as information retriever (IR) to  collect candidate paragraphs.  Pas EM: exact match of gold paragraphs; Ans EM/F1: exact match/F1 on short answer; Sup EM/F1: exact match/F1 on supporting facts. }
\label{tbl:hotpot}
\end{table*}

\subsection{Datasets}
We conduct experiments on five datasets, the statistics of which is shown in Table~\ref{tbl:stat}.

\textbf{MNLI}~\cite{N18-1101}\footnote{https://gluebenchmark.com/tasks}: Multi-Genre Natural Language Inference matched (MNLI-m) and mismatched (MNLI-mm) are textual entailment tasks. The goal is to classify the relation between premise and hypothesis sentences into three classes: entailment, contradiction and neutral. The train and test sets come from the same \emph{source} and same \emph{genre} in MNLI-m, and different in MNLI-mm.

\textbf{SNLI}~\cite{bowman:emnlp15}\footnote{https://nlp.stanford.edu/projects/snli/}: 
The dataset of Stanford Natual Language Inference is another textual entailment task.

\textbf{QQP}~\cite{wang2018glue}: Quora Question Pairs is to identify whether two questions are duplicated or not.

\textbf{Quasar-T}~\cite{dhingra2017quasar}\footnote{https://github.com/bdhingra/quasar}: This is a dataset for question answering by searching the related passages and then reading it to extract the answer.
In this dataset, we evaluate the models by whether it can correctly select the sentence containing the gold answer from the candidate pool.

\textbf{HotpotQA}~\cite{yang2018hotpotqa}\footnote{https://hotpotqa.github.io/}: A dataset of diverse and explainable multi-hop question answering.
We focus on the full-wiki setting, where the model needs to extract the answer from all the abstracts in Wikipedia and related sentences.

Note that for the datasets of MNLI, QQP and HotpotQA, the test sets are hidden and the number of submissions is limited.
For a fair comparison between our models and baselines, we split 5\% of the training data as validation set and use the original validation set as test set.

\subsection{Implementation Details}
All the models are pre-trained on Wikipedia and finetuned on downstream tasks.
We also evaluate the pre-trained models on whether they can perform unsupervised paragraph selection, and whether the improvement over paragraph ranking can lead to better answer prediction on HotpotQA task.
As our experiments are to evaluate the ability of sentence encoder, we only build a light layer on sentence embeddings for classification task, and use only dot product (Cross-Thought, ICT) or cosine similarity (Skip-thought, LM, MLM) between sentence embeddings for ranking task. 
Note that for fair comparison, all the encoders in our experiments have the same structure as RoBERTa-base (12 layers, 12 self-attention heads, hidden size 768). 
For all experiments, we use Adam~\cite{kingma2014adam} as the optimizer and use the tokenizer of GPT-2~\cite{gpt}.

For model pre-training, all models including the baselines we re-implement are trained with Wikipedia pages.\footnote{The Wikipedia dump we use is enwiki-20191001-pages-articles-multistream.}
We use 16 NVIDIA V100 GPUs for model training.
Our code is mainly based on the RoBERTa codebase,\footnote{https://github.com/pytorch/fairseq} and we use similar hyper-parameters as RoBERTa-base training.
Each training sample contains 500 short sequences with 64 tokens, and we randomly mask 15\% of the tokens in the sequences.
During training, we fix the position embeddings for the pre-appended special tokens, and randomly select 64 continuous positions from 0 to 564 for the other words.
Thus, the model can be used to encode longer sequences in downstream tasks.
The batch size is set to 128 (4 million tokens).
We use warm-up steps 10,000, maximal update steps 125,000, learning rate 0.0005, dropout 0.1 for pre-training.
Each model is pre-trained for around 4 days.

For model finetuning, in experiments for MNLI, SNLI and QQP, we use batch size 32, warmup steps 7,432, maximal update steps 123,873, and learning rate 0.00001. 
For Quasar-T and HotpotQA, we set batch size 80, warmup steps 2,220, maximal update steps 20,360, and learning rate 0.00005.
Dropout is the only hyper-parameter we tuned, and 0.1 is the best from [0.1, 0.2, 0.3].
As HotpotQA in full-wiki setting does not provide answer candidates, we randomly sample 100 negative paragraphs from the top 1000 paragraphs ranked by BM25 scores during training.
During inference, we use sentence embeddings to further rank the top 1000 paragraphs.
For the unsupervised experiment on HotpotQA, we only rank top 200 paragraphs.

\subsection{Baselines}
Existing baseline methods are mostly trained with different encoders or different datasets.
For fair comparison, we re-implement all these baselines by using a 12-layer Transformer as the sentence encoder and Wikipedia as the source for pre-training. There are three groups of baselines considered for evaluation:

\paragraph{Pre-trained Sentence Embedding}

\begin{itemize}[itemsep=1pt,topsep=2pt,leftmargin=12pt]
  \item \textit{ICT}~\cite{ict}: Inverse cloze task treats a sentence and its context as a positive pair, otherwise negative. 
Sentences are masked from the context 10\% of the time.
This model is trained by ranking loss based on the dot product between sequence embeddings.

 \item \textit{Skip-Thought}~\cite{skipthought}: The task is to encode sentences into embeddings that are used to re-construct the next and the previous sentences. This model is based on encoder-decoder structure without considering attention across sequences~\cite{cho2014learning}. We use 6-layer Transformer as the decoder for re-construction. 
\end{itemize}

\paragraph{Language Modeling}

In addition, we also re-implement benchmark baselines on the classic Language Modeling (LM) and Mask Language Modeling tasks, as most existing models are pre-trained with different unlabeled datasets:

\begin{itemize}[itemsep=1pt,topsep=2pt,leftmargin=12pt]
  \item \textit{Language Model (LM)}~\cite{gpt}: The task is to predict the probability of the next word based on given context. As the words are sequentially encoded, to evaluate the performance of this model on HotpotQA in the unsupervised setting, we use the last hidden state as the sentence embedding (instead of the first one by ICT, MLM, Skip-Thought).

\item \textit{Masked Language Model (MLM)}~\cite{bert}: The task is to generate randomly masked words from sequences. We explore different settings of training data. ``Masked LM-1-512" trains a Transformer on sequences with 512 tokens and pre-appends 1 special token to each sequence. ``Masked LM-1-64" is trained on sequences with 64 tokens. Both models are trained with Wikipedia text only. ``Masked LM (160G)" is the RoBERTa model pre-trained on a much larger corpus.
\end{itemize}

\begin{table*}[]
\small
\begin{tabular}{lp{7cm}p{7cm}}
\toprule
Q & Jens Risom introduced what type of design, characterized by minimalism and functionality?          & Washington was named after President \textless{}\textcolor{red}{MASK}\textgreater by an act of the \textless{}\textcolor{blue}{MASK}\textgreater Congress during the creation of Washington Territory in 1853.              \\
\midrule
C1    & \textcolor{blue}{\textbf{Scandinavian design}} is a design movement characterized by simplicity, minimalism and  functionality that emerged in the 1950s in the five Nordic countries of Finland, Norway,  Sweden, Iceland and Denmark...  (attention weight: \textbf{0.259}) 
& Washington, officially the State of Washington, is a state in the Pacific Northwest region of the \textcolor{blue}{United States}.  Named for \textcolor{red}{\textbf{George Washington}}, the first U.S. president. (attention weight: \textbf{0.987})                                  \\
C2    & Risom was one of the first designers to introduce \textcolor{blue}{\textbf{Scandinavian design}} in the United States...  (attention weight: \textbf{0.194}) 
& Approximately 60 percent of Washington‘s residents live in the Seattle metropolitan area,  the center of transportation, business, and industry. (attention weight: \textbf{0.012})                                                       \\
C3    & Dutch Design can be characterized as minimalist, experimental, innovative, quirky, and humorous...  (attention weight: \textbf{0.098})                                                                             & Manufacturing industries in Washington include aircraft and missiles, shipbuilding, and other ... (attention weight: \textbf{0.001})\\
\bottomrule
\end{tabular}
\caption{Case study on unsupervised passage ranking. The attention weights are learned by cross-sequence Transformer from pre-training. The examples on the left  come from HotpotQA and are the ranked passages from 200 candidates for answering the question. The examples on the right are in the format of Masked Language Modeling task, where our Cross-Thought needs to recover the masked words by leveraging other sequences. C: the ranked passages by attention weights. }
\label{tbl:case}
\end{table*}

\paragraph{Multi-hop Question Answering}

To further evaluate our model on multi-hop question answering task in open-domain setting, we compare our framework with several strong baselines on HotpotQA:

\begin{itemize}[itemsep=1pt,topsep=2pt,leftmargin=12pt]
  \item \textit{Cognitive Graph}~\cite{ding2019cognitive}: It uses an iterative process of answer extraction and further reasoning over graphs built upon extracted answers.

\item \textit{Semantic Retrieval}~\cite{nie2019revealing}: It uses a semantic retriever on both paragraph- and sentence-level to retrieve question-related information.

\item  \textit{Recurrent Retriever}~\cite{asai2019learning}: It uses a recurrent retriever to collect useful information from Wikipedia graphs for question answering.
\end{itemize}

\subsection{Experimental Results}
Results on the classification and ranking tasks are summarized in Table~\ref{tbl:exp}. Results of our pipeline on HotpotQA (full-wiki) are shown in Table~\ref{tbl:hotpot}.

\paragraph{Effect of Pre-training Tasks} 
Among all the pre-training tasks, our proposed method Cross-Thought achieves the best performance. 
With finetuning, LM pre-training tasks work better than the Skip-Thought and ICT methods which are specifically designed for learning sentence embedding. 
Moreover, we provide a fair comparison between ``Cross-thought-1-64" and ``Masked LM-1-64", both of which segment Wikipedia text into short sequences in 64 tokens for pre-training, and only use the hidden state of the first special token as sentence embedding.
Results show that our Cross-Thought model  achieves much better performance than Masked LM-1-64, as well as the Transformer pre-trained on 160G data (10 times larger than Wikipedia).

\paragraph{Effect of Training on Short Sequences}
Results on ``Cross-Thought-1-512" and ``Cross-Thought-1-64" (using sequences of 512 tokens and 64 tokens, respectively) clearly show that shorter sequences lead to more effective pre-training. 
Moreover, we also observe that ``Cross-Thought-1-512" and ``Masked LM-1-512" achieve almost the same performance.
It means that our Cross-Thought has to be trained on short sequences (64 tokens); otherwise, it would learn more on the word dependencies within sequence other than the sequence embeddings.
Actually, the effect of short sequences is also proved by Skip-Thought which focuses on generating sequences in sentence level, but our Cross-Thought can achieve better performance.

\paragraph{Effect of Sentence Embedding Size}
As we keep the number of parameters fixed for the encoders trained with different tasks, increasing the dimension of hidden state will lead to more parameters to train. Instead, for each sequence, we pre-append more special tokens, the hidden states of which are concatenated together as the final sentence embedding. 
Experiments on ``Cross-Thought-1-64", ``Cross-Thought-3-64" and ``Cross-Thought-5-64" compare pre-appending 1, 3 and 5 different special tokens to sequences for pre-training.
We can see that a larger sentence embedding size can significantly improve performance on the ranking tasks while not on the classification tasks.
We hypothesize that the main reason is ranking tasks are more challenging, with many different pairs to compare, for which the contextual sentence embeddings can provide additional information.

\paragraph{Effect of Paragraph Ranking without Finetuning}
We also conduct an analysis on whether pre-trained sentence embeddings can be directly used for downstream tasks without finetuning.
Although model performance without finetuning is generally worse than supervised training, experiments in column ``HotpotQA(u)'' further validate the previously discussed three conclusions.
Besides, we observe that although the model pre-trained by masked language modeling leads to better performance after finetuning, it is not designed to train sentence embeddings, thus cannot be used for passage ranking. 
While all the other methods achieve much better performance than masked language modeling,
our model ``Cross-Thought-5-64" with the largest embedding size achieves the best performance.

\paragraph{Effect of Cross-Thought as Information Retriever (IR) on QA Task}
Our pipeline of solving HotpotQA (full-wiki) consists of three steps: $(i)$ Fast candidate paragraph retrieval; $(ii)$ Multi-hop paragraphs re-ranking by a more complex model; and $(iii)$ Answer and supporting facts extraction.
We evaluate our proposed method on how well the finetuned sentence embeddings can be utilized in the first step for IR, with the re-ranker and answer extractor fixed. ``Masked LM-1-64" and ``Cross-Thought-1-64" in Table~\ref{tbl:hotpot} show that our pre-trained model achieves better performance than the baseline model pre-trained on single sequences.
Moreover, the pipeline integrating our sentence embedding achieves new state of the art on HotpotQA (full-wiki).

\subsection{Case Study}
Table~\ref{tbl:case} provides a case study on the unsupervised passage ranking and masked language modeling tasks.
For the case from HotpotQA, we can see that attention weights from the cross-sequence Transformer in Cross-Thought can rank the paragraph with gold answer to the first place among the 200 candidate paragraphs.

For the case from Mased Language Modeling, we also observe that the sentence that can be used to recover the masked words receives much higher attention weight compared to others, validating our motivation on retrieving the useful sentence embeddings from other sequences to enhance masked word recovery in the current sequence. 

\section{Conclusion}
We propose a novel approach, Cross-Thought, to pre-train sentence encoder. Experiments demonstrate that using Cross-Thought trained with short sequences can effectively improve sentence embedding. Our pre-trained sentence encoder with further finetuning can beat several strong baselines on many NLP tasks. 
\bibliographystyle{acl_natbib}
\bibliography{emnlp2020}

  
\end{document}